\documentclass[]{fairmeta}
\usepackage{makecell}
\usepackage{wrapfig}
\usepackage{tabularx}
\usepackage{textcomp}
\usepackage{stfloats}
\usepackage{url}
\usepackage{verbatim}
\usepackage{titlesec}
\usepackage{tocloft}
\usepackage{adjustbox}
\usepackage{multirow}
\usepackage{pifont}
\usepackage[sc]{mathpazo}
\usepackage{tikz}
\usepackage{comment}
\usepackage{amsmath,amssymb}
\usepackage{colortbl}
\usepackage{natbib}
\usepackage{color}
\usepackage{booktabs} 
\usepackage{hyperref}
\usepackage{graphicx}
\usepackage{subcaption}
\RequirePackage{xspace}
\makeatletter
\DeclareRobustCommand\onedot{\futurelet\@let@token\@onedot}
\def\@onedot{\ifx\@let@token.\else.\null\fi\xspace}
\usepackage[most]{tcolorbox}
\usepackage{array}
\usepackage{siunitx}
\usepackage[table]{xcolor}
\usepackage{caption}
\definecolor{headerpurple}{HTML}{d8d2fc}
\definecolor{rowgray}{gray}{0.95}
\usepackage{CJKutf8}

\makeatother

\definecolor{adptorange}{RGB}{248, 205, 172}
\definecolor{cmpblue}{RGB}{189, 215, 238}

\definecolor{our_red}{RGB}{232,157,160}
\definecolor{our_blue}{RGB}{136,206,230}
\definecolor{our_orange}{RGB}{246,200,168}
\definecolor{our_green}{RGB}{178,211,164}

\definecolor{attn_code0}{RGB}{247,215,200}
\definecolor{attn_code1}{RGB}{238,169,139}
\definecolor{mlp_code0}{RGB}{204,201,221}
\definecolor{mlp_code1}{RGB}{102,95,153}
\definecolor{mygray}{HTML}{f0f0f0}

\definecolor{token_blue}{RGB}{84, 120, 140}

\usepackage{bbding}
\usepackage{fontawesome}
\usepackage{float}

\newlength\savewidth

\newcolumntype{x}[1]{>{\centering\arraybackslash}p{#1pt}}
\newcolumntype{y}[1]{>{\raggedright\arraybackslash}p{#1pt}}
\newcolumntype{z}[1]{>{\raggedleft\arraybackslash}p{#1pt}}

\renewcommand{\paragraph}[1]{\vspace{1.25mm}\noindent\textbf{#1}}

\usepackage{algorithm}
\usepackage{listings}

\definecolor{codeblue}{rgb}{0.25, 0.5, 0.5}
\definecolor{codekw}{rgb}{0.35, 0.35, 0.75}
\lstdefinestyle{Pytorch}{
    language = Python,
    backgroundcolor = \color{white},
    basicstyle = \fontsize{9pt}{8pt}\selectfont\ttfamily\bfseries,
    columns = fullflexible,
    aboveskip=1pt,
    belowskip=1pt,
    breaklines = true,
    captionpos = b,
    commentstyle = \color{codeblue},
    keywordstyle = \color{codekw},
}

\definecolor{green}{HTML}{009000}
\definecolor{red}{HTML}{ea4335}

\title{AlayaWorld: Interactive Long-Horizon World Modeling - Full Technical Report (v1.1)}
\author{AlayaWorld Team, Alaya Lab *}
\abstract{
This report presents an improved version of AlayaWorld. While the backbone architecture, chunk-wise autoregressive generation scheme, and training data remain unchanged from the previous release, we substantially revise how conditioning signals are represented and integrated into the model.
The new design is guided by a simple principle: conditioning signals should match the generated content as closely as possible in both latent representation and temporal structure. To this end, we make two major changes. First, we replace the previous depth-warping-based spatial memory with a streaming 3D point-cache renderer. Second, we redesign the conditioning pipeline so that visual conditions are encoded in the same causal-VAE latent space, with temporal statistics consistent with those of the generated video.
Concretely, the new version introduces six modifications: (1) replacing static-frame image conditioning with motion-aware latent conditioning; (2) causally encoding re-rendered spatial memory as a continuous sequence; (3) aligning the temporal-memory window in pixel space; (4) adopting hard memory dropout that removes memory tokens rather than zeroing them; (5) unifying the VAE encoding and decoding protocol across training and inference; and (6) removing the camera AdaLN branch, such that viewpoint control is provided entirely through the re-rendered spatial condition.

}

\github{\url{https://alaya-lab.github.io/AlayaWorld/}}
\Code{\url{https://github.com/AlayaLab/AlayaWorld}}
\video{\url{https://www.youtube.com/watch?v=n0jIEg7taTI}}

\contact{kaipeng.zhang@shanda.com}
\date{\today}

\begin{document}
\maketitle

\begingroup
\renewcommand{\thefootnote}{}
\footnotetext{* Alaya Lab contributors are listed at the end of the report.}
\endgroup

\section{Conditioning and Memory Redesign}
\label{sec:changes}

The previous version of AlayaWorld conditioned generation on text, temporal memory, spatial memory, and an explicit camera-control branch.
While these signals provided complementary information, they were processed through different pathways: visual conditions were often encoded independently, temporal memory was not always aligned with the causal structure of the VAE, and camera trajectories were injected separately through AdaLN.
As a result, the conditioning signals seen by the model could differ from the generated video latents in their temporal structure, latent representation, or train--inference behavior.

The new version redesigns the conditioning pipeline around two principles.
First, visual conditions should follow the same causal-VAE encoding structure as the video being generated.
Second, viewpoint control should be expressed through geometry-aligned visual conditions rather than through a separate modulation branch.
Based on these principles, we introduce the following six changes.

\paragraph{1. Motion-aware image conditioning.}
In the previous version, an image condition was encoded as an isolated frame, yielding a latent representation without the temporal context present in causally encoded video latents.
We instead construct a nine-frame window consisting of the conditioning frame and its preceding eight frames ($\text{stride}+1=9$), encode the full window with the causal VAE, and use the second latent as the image condition.
The single conditioning frame is retained only as a decoder prefix.
The same nine-frame encoding pattern is used for chunk-to-chunk handoff during inference, so image conditioning and autoregressive continuation are represented with matched temporal context.

\paragraph{2. Causal encoding of a streaming 3D spatial memory.}
We replace the previous DA3-based depth warping \cite{depthanything3} with a streaming 3D point cache, using ViGeo \cite{yu2026vigeo} to estimate the per-pixel 3D geometry.
After each generated chunk, per-pixel 3D points are registered into a persistent cache, which is then re-rendered from the planned viewpoint of the next chunk.
To maintain geometric consistency over the rollout, the camera trajectory is scale-aligned to the cache using a pairwise-median ratio over prefix displacements, while camera intrinsics are estimated once per clip by least-squares fitting of a pinhole camera model to the point map and then kept fixed.
We also change how the rendered spatial condition is encoded.
Instead of encoding rendered frames independently, we prepend a real prefix frame and encode the resulting sequence causally.
The prefix latent is then discarded, leaving spatial-memory latents that follow the same causal encoding structure as the target video latents.
Tokens corresponding to invalid rendered regions are removed rather than masked, preserving compatibility with FlashAttention.

\paragraph{3. Pixel-aligned temporal memory.}
Temporal memory is reduced from six latents to four and is constructed by aligning the memory window in pixel space before VAE encoding.
For $N=4$ memory latents, we encode exactly $1+(N-1)\times 8=25$ frames, producing exactly four causal latents.
The data loader and trainer use the same frame-to-latent convention to ensure consistent window boundaries.
This removes the off-by-one boundary latents that could previously leak into the temporal-memory context.

\paragraph{4. Hard memory dropout.}
The previous memory-dropout strategy zeroed memory tokens while preserving their positions in the sequence.
Although their values were removed, the sequence structure still revealed whether memory tokens were present.
We replace this with hard dropout, which removes the memory tokens entirely and therefore changes the actual sequence length.
Correspondingly, the first rollout step at inference is performed without temporal memory, matching the memory-free cases seen during training.

\paragraph{5. A unified VAE protocol across training and inference.}
We standardize the pixel--latent interface used throughout training, autoregressive rollout, and evaluation.
Chunk handoff consistently uses an anchor frame together with the same nine-frame causal encoding window.
Decoding supports both direct latent continuation and an RGB decode--re-encode path, while evaluation supplies ground-truth continuations together with the two causal prefix latents required by the VAE.
This removes several stage-specific encoding and decoding conventions from the previous version and reduces discrepancies between training and autoregressive inference.

\paragraph{6. Geometry-based camera control.}
The dedicated camera AdaLN branch is removed.
Instead, the planned camera trajectory is used to determine the viewpoint from which the 3D point cache is re-rendered, and the resulting spatial condition provides the camera guidance to the model.
After causal-VAE encoding, viewpoint information is therefore presented through the same visual latent representation as other spatial conditions.
This directl
y couples camera control to scene geometry, including scale, visibility, and parallax, without requiring a separate camera-conditioning side channel.
Text conditioning continues to specify scene semantics, while the rendered spatial memory specifies the desired viewpoint.
\section{Experimental Results}

\subsection{Quantitative Results}

\newlength{\thickarrayrulewidth}
\setlength{\thickarrayrulewidth}{1pt}
\newcommand{\thickhline}{\noalign{\hrule height \thickarrayrulewidth}}

\begin{table*}[t]
\caption{
            Quantitative comparison across all scenarios on WBench navigation split (158 navigation cases).
            All scores $\in [0,100]$, higher is better.
            \textbf{Bold} = best, \underline{underline} = second best.
        }
\label{tab:wbench}
\centering
\renewcommand{\arraystretch}{1.3}
\setlength{\tabcolsep}{2.33pt}
\scriptsize
\resizebox{\textwidth}{!}{%
\begin{tabular}{@{}l *{9}{c} @{}}
\thickhline
\textbf{Metrics}
  & \scriptsize\textbf{Yume 1.5}
  & \scriptsize\textbf{Matrix-Game 2.0}
  & \scriptsize\textbf{HY-World 1.5}
  & \scriptsize\textbf{HY-GameCraft}
  & \scriptsize\textbf{LingBot-Fast}
  & \scriptsize\textbf{LingBot-v2}
  & \scriptsize\textbf{Genie 3}
  & \scriptsize\textbf{Happy Oyster}
  & \scriptsize\textbf{AlayaWorld} \\
\hline
\multicolumn{10}{@{}l}{\textbf{Video Quality}} \\
  \quad Aesthetic   & 58.7 & 54.0 & 60.1 & 52.6 & 62.6 & \textbf{64.4} & 51.6 & 56.6 & \underline{62.8} \\
  \quad Imaging     & 63.3 & 60.3 & 65.4 & 58.7 & 63.8 & \underline{67.5} & 59.3 & 63.9 & \textbf{67.7} \\
  \quad Flickering  & 93.0 & \underline{94.6} & 93.5 & 93.7 & 92.4 & 91.4 & \textbf{95.0} & 94.0 & 92.7 \\
  \quad Dynamic     & \textbf{96.8} & 94.9 & 91.1 & \textbf{96.8} & 95.6 & \underline{96.2} & 92.4 & 94.2 & 91.1 \\
  \quad Smoothness  & 97.0 & \textbf{98.2} & \underline{98.1} & 97.6 & 96.0 & 96.5 & 97.8 & 97.0 & 97.0 \\
  \quad HPSv3-Norm  & 57.0 & 41.0 & 60.5 & 38.3 & \underline{65.7} & \textbf{74.6} & 55.2 & 58.3 & 64.3 \\
  \quad \textit{Average} & 77.6 & 73.8 & 78.1 & 73.0 & \underline{79.4} & \textbf{81.8} & 75.2 & 77.3 & 79.3 \\
\hline
\multicolumn{10}{@{}l}{\textbf{Setting}} \\
  \quad Scene       & 53.1 & 49.4 & 53.5 & 50.6 & \underline{63.4} & \textbf{66.7} & 61.1 & 57.4 & 51.6 \\
  \quad Subject     & \underline{91.7} & 84.9 & 90.8 & 82.5 & \textbf{92.4} & 86.9 & 83.8 & 91.1 & 87.7 \\
  \quad \textit{Average} & 72.4 & 67.2 & 72.2 & 66.6 & \textbf{77.9} & \underline{76.8} & 72.5 & 74.3 & 69.7 \\
\hline
\multicolumn{10}{@{}l}{\textbf{Interaction}} \\
  \quad Navigation  & 72.0 & 80.6 & \textbf{87.5} & 67.8 & 79.4 & 82.8 & 73.3 & \underline{85.1} & 80.0 \\
\hline
\multicolumn{10}{@{}l}{\textbf{Consistency}} \\
  \quad Background    & 90.3 & 86.9 & \underline{92.7} & 86.5 & 90.9 & 92.5 & 90.7 & 91.4 & \textbf{94.1} \\
  \quad Spatial       & 71.5 & 64.5 & \textbf{90.6} & 60.5 & 77.2 & 82.3 & 79.9 & 77.7 & \underline{87.9} \\
  \quad Gated Spatial & 71.4 & 64.5 & \textbf{84.9} & 60.5 & 76.9 & 78.7 & 78.4 & 75.8 & \underline{81.9} \\
  \quad Segment       & \underline{99.4} & 21.0 & \textbf{100.0} & \underline{99.4} & 98.1 & 98.1 & 93.6 & 96.2 & 98.1 \\
  \quad Perspective   & 48.0 & 29.2 & 62.5 & 17.9 & 82.8 & \underline{84.5} & 54.5 & 75.0 & \textbf{86.6} \\
  \quad Subject       & 88.8 & 87.2 & 89.1 & 82.6 & 88.6 & 88.9 & 90.4 & \underline{91.5} & \textbf{93.4} \\
  \quad Geometric     & 88.0 & 86.1 & \underline{92.0} & 88.3 & 85.4 & 87.1 & 88.6 & 87.2 & \textbf{94.1} \\
  \quad Photometric   & 83.3 & 81.3 & 83.1 & \textbf{85.0} & 79.1 & 79.8 & \underline{84.5} & 79.8 & 80.3 \\
  \quad \textit{Average} & 80.1 & 65.1 & \underline{86.9} & 72.6 & 84.9 & 86.5 & 82.6 & 84.3 & \textbf{89.5} \\
\hline
\multicolumn{10}{@{}l}{\textbf{Physical}} \\
  \quad Causal Fidelity     & 72.7 & 59.3 & \underline{74.0} & 68.3 & 72.5 & \textbf{76.7} & 71.7 & 69.3 & 65.1 \\
  \quad Visual Plausibility & 57.7 & 55.0 & 58.6 & 56.5 & 58.8 & \textbf{61.4} & 59.7 & 57.6 & \underline{61.1} \\
  \quad \textit{Average}    & 65.2 & 57.2 & \underline{66.3} & 62.4 & 65.7 & \textbf{69.1} & 65.7 & 63.5 & 63.1 \\
\thickhline
\end{tabular}
}
\end{table*}

Table~\ref{tab:wbench} reports the comparison on the WBench \cite{ying2026wbench} navigation split, covering 158 navigation cases across Video Quality, Setting, Interaction, Consistency, and Physical metrics. AlayaWorld performs strongly on the Consistency benchmarks, achieving the best overall Consistency score of 89.5. It also remains competitive in Video Quality, with an average score of 79.1, while maintaining solid navigation performance under the interactive evaluation protocol. Overall, the results indicate that AlayaWorld is effective at preserving visual and geometric consistency during long-horizon interactive generation.

For Video Quality, AlayaWorld achieves the best Imaging score of 67.7 and remains competitive in Aesthetic quality with a score of 62.6. Its overall Video Quality score of 79.1 is close to the strongest competing methods, despite not leading on temporal metrics such as Flickering, Dynamic, and Smoothness. These results suggest that AlayaWorld maintains high perceptual quality and image fidelity during autoregressive interaction, while leaving room for further improvement in short-term temporal dynamics.

The main advantage of AlayaWorld is observed in Consistency. It achieves the best results in Background Consistency, Perspective Consistency, Subject Consistency, and Geometric Consistency, and ranks second in both Spatial and Gated Spatial Consistency. In particular, the strong Perspective and Geometric scores indicate that AlayaWorld better preserves scene structure and viewpoint-dependent geometry as the camera moves through the environment, while the high Background and Subject scores show that previously observed visual content remains stable over extended interactions. Together, these improvements lead to the highest overall Consistency score among all evaluated methods, validating the effectiveness of the proposed spatial and temporal memory mechanisms for long-horizon generation.

For Interaction, AlayaWorld achieves a Navigation score of 79.9, demonstrating responsiveness to navigation controls, although competing methods perform better.
Its performance on the Setting and Physical categories is comparatively weaker, particularly for Scene consistency and Causal Fidelity.
These results suggest that while AlayaWorld provides strong visual persistence and geometric stability, improving environment-level semantic preservation and physical interaction modeling remains an important future direction.

\subsection{Qualitative Results}

\begin{figure}[htbp]
    \centering
    \includegraphics[width=\linewidth]{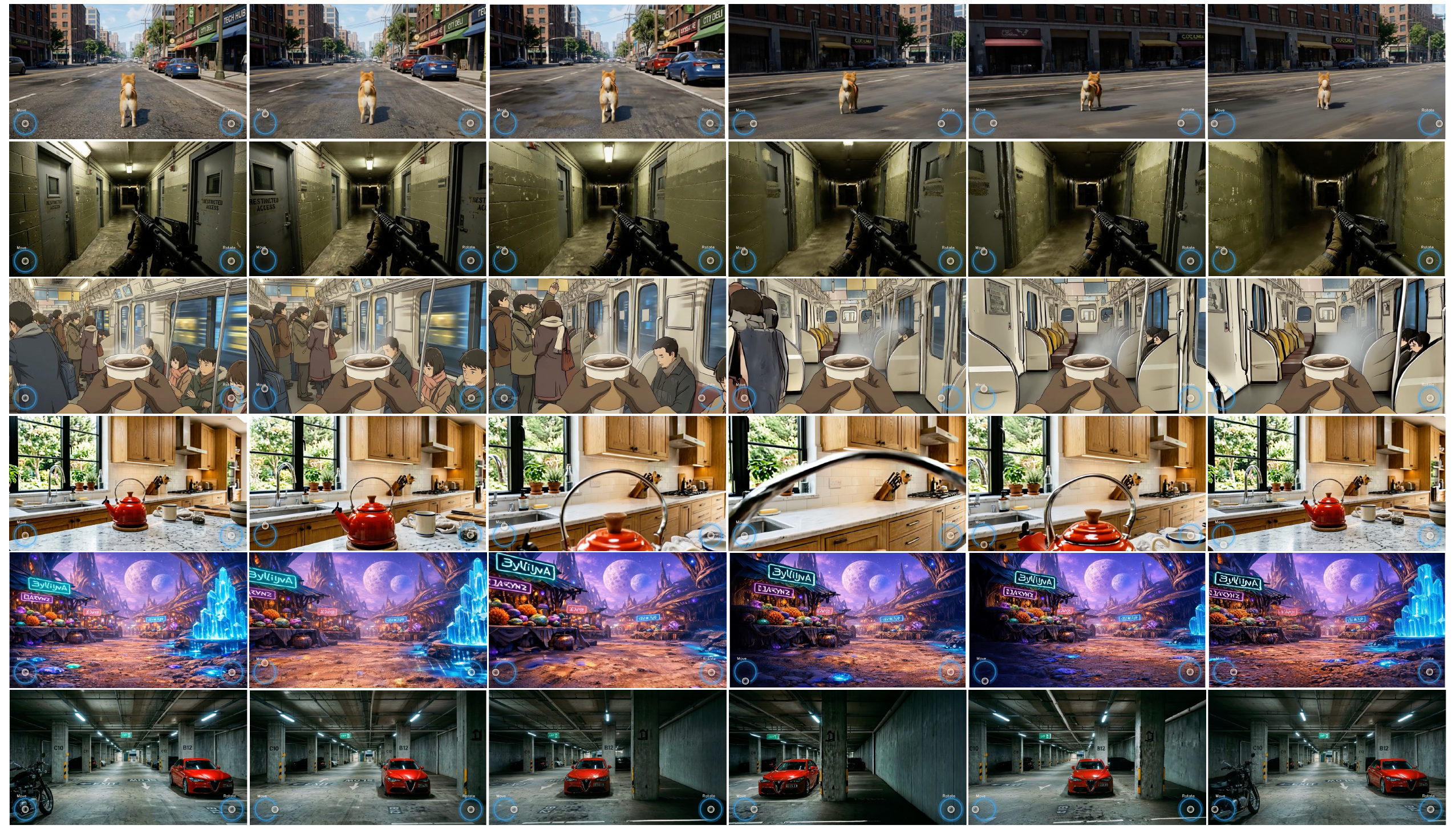}
    \caption{
    Qualitative results of AlayaWorld on WBench \cite{ying2026wbench}.
    }
    \label{fig:wbench}
\end{figure}

\begin{figure}[htbp]
    \centering
    \includegraphics[width=\linewidth]{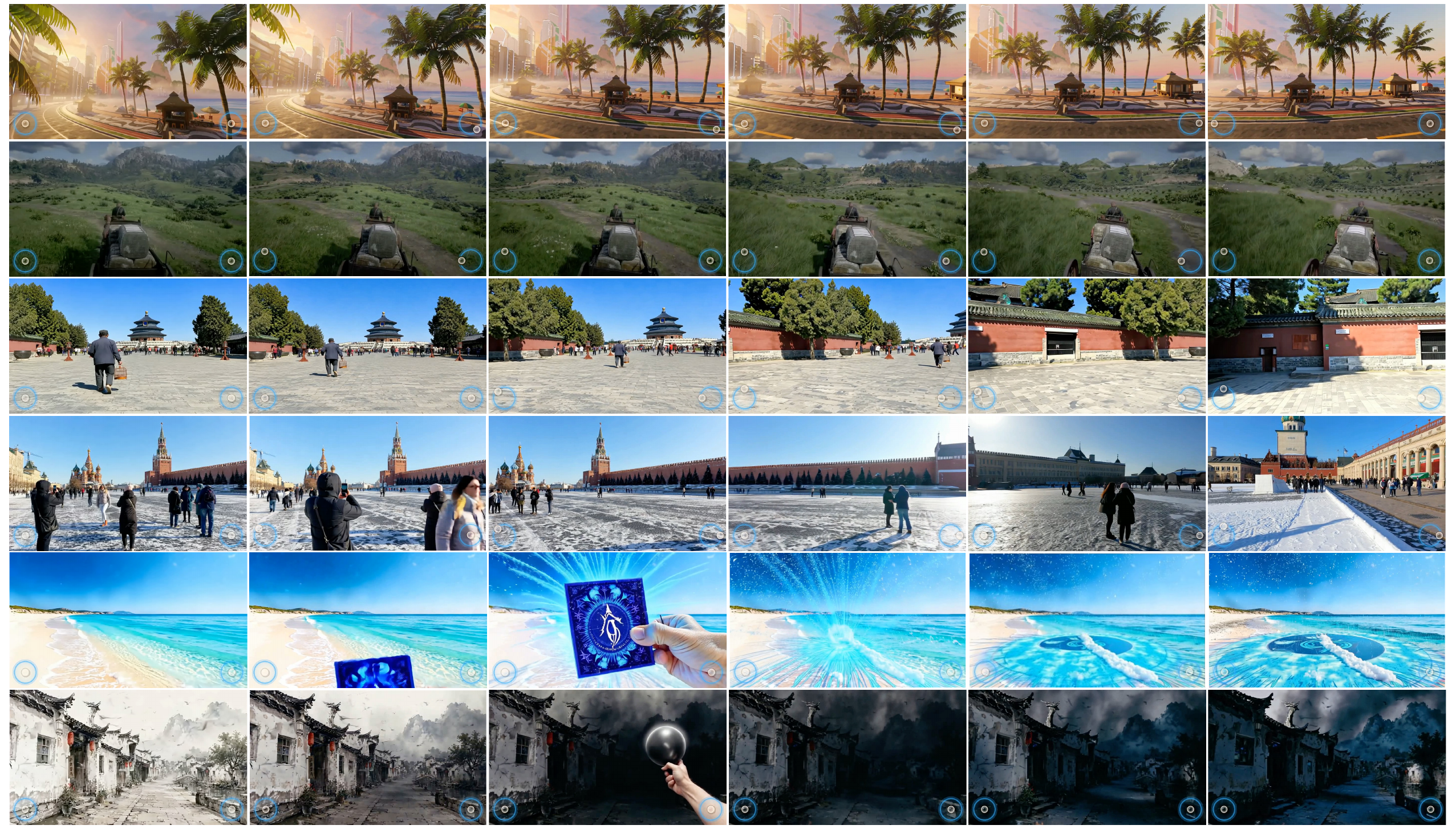}
    \caption{
    Qualitative results of AlayaWorld across different scenes.
    }
    \label{fig:qual}
\end{figure}

Figure~\ref{fig:wbench} presents qualitative results of AlayaWorld on WBench.
Across diverse navigation scenarios, AlayaWorld produces coherent visual transitions while maintaining the appearance of scene content and the spatial relationships among objects throughout the interaction.
The results further show stable scene structure under continuous viewpoint changes, demonstrating strong visual persistence and spatial consistency during interactive navigation.
Figure~\ref{fig:qual} presents qualitative results of AlayaWorld across a diverse set of scenes.
The results demonstrate that the model generalizes well across different environments and maintains consistent visual content under varying scene layouts and motion patterns.

\section{Contributions and Acknowledgments}

Within each role category, authors are listed in alphabetical order by their first names.

\textbf{Core Lead:}
Kaipeng Zhang

\textbf{Lead:}
Chuanhao Li

\textbf{Core Contributor:}
Chuanhao Li, Kaipeng Zhang, Yifan Zhan, Yongtao Ge, Yuanyang Yin

\textbf{Contributor:}
Jiaming Tan, Kang He, Liaoyuan Fan, Mingliang Zhai, Ruicong Liu, Xiaojie Xu, Xuangeng Chu, Zhen Li, Zhengyuan Lin, Zhixiang Wang, Zian Meng, Zihui Gao

\bibliographystyle{abbrv}
\bibliography{references}

@article{ying2026wbench,
  title={WBench: A Comprehensive Multi-turn Benchmark for Interactive Video World Model Evaluation},
  author={Ying, Kaining and Hu, Hengrui and Ren, Siyu and Li, Jiamu and Chen, Fengjiao and Wang, Ziwen and Cao, Xuezhi and Cai, Xunliang and Ding, Henghui},
  journal={arXiv preprint arXiv:2605.25874},
  year={2026}
}

@article{yu2026vigeo,
  title={Towards Consistent Video Geometry Estimation},
  author={Yu, Zhu and Gao, Jingnan and Zhang, Runmin and Qiu, Lingteng and Zhao, Zhengyi and Peng, Rui
          and Yan, Yichao and Qiu, Kejie and Zhu, Siyu and Dong, Zilong and Cao, Si-Yuan and Shen, Hui-Liang},
  journal={arXiv:2605.30060},
  year={2026}
}

@article{depthanything3,
  title={Depth Anything 3: Recovering the visual space from any views},
  author={Haotong Lin and Sili Chen and Jun Hao Liew and Donny Y. Chen and Zhenyu Li and Guang Shi and Jiashi Feng and Bingyi Kang},
  journal={arXiv preprint arXiv:2511.10647},
  year={2025}
}

\end{document}